\documentclass[conference]{IEEEtran}
\IEEEoverridecommandlockouts
\usepackage{cite}
\usepackage{amsmath,amssymb,amsfonts}
\usepackage{graphicx}
\usepackage{textcomp}

\usepackage[table]{xcolor}
\usepackage{times}
\usepackage{soul}
\usepackage{url}
\usepackage[hidelinks]{hyperref}
\usepackage[utf8]{inputenc}
\usepackage[small]{caption}
\usepackage{tablefootnote}
\usepackage{graphicx}
\usepackage{amsthm}
\usepackage{amsmath,amssymb,amsfonts}
\usepackage{booktabs}
\usepackage{algorithm}
\usepackage{algorithmicx}
\usepackage{algpseudocode}
\usepackage[normalem]{ulem}

\usepackage{subcaption}
\useunder{\uline}{\ul}{}
\def\BibTeX{{\rm B\kern-.05em{\sc i\kern-.025em b}\kern-.08em
    T\kern-.1667em\lower.7ex\hbox{E}\kern-.125emX}}
\begin{document}

\title{Scalable Counterfactual Explanations for Time Series via Policy Learning}

\author{
\IEEEauthorblockN{
Peiyu Li\IEEEauthorrefmark{1},
Pouya Hosseinzadeh\IEEEauthorrefmark{2},
Souka\"ina Filali Boubrahimi\IEEEauthorrefmark{2},
Shah Muhammad Hamdi\IEEEauthorrefmark{2}
}

\IEEEauthorblockA{\IEEEauthorrefmark{1}
McComish Department of Electrical Engineering and Computer Science,\\
South Dakota State University, Brookings, SD, USA\\
Email: \texttt{peiyu.li@sdstate.edu}
}

\IEEEauthorblockA{\IEEEauthorrefmark{2}
Department of Computer Science,\\
Utah State University, Logan, UT, USA\\
Email: \texttt{\{pouya.hosseinzadeh, soukaina.boubrahimi, s.hamdi\}@usu.edu}
}
}

\maketitle

\begin{abstract}
Generating counterfactual (CF) explanations for time series is challenging due to complex temporal dependencies and the computational inefficiency of traditional instance-specific optimization. We propose a policy learning framework for counterfactual explanation in time series, trained as a one-step contextual-bandit policy-learning problem. By training an actor-critic agent in the latent space of a pre-trained autoencoder, our method learns a transferable policy that captures global structural regularities. Unlike existing methods that rely on iterative optimization for each new input, the learned policy enables direct generation of valid and plausible counterfactuals through a single forward pass, avoiding per-instance optimization at inference time. Experiments on 13 UCR benchmark datasets show that our framework achieves high validity and strong plausibility across multiple anomaly detection metrics. Overall, by decoupling counterfactual generation from per-instance optimization, the proposed method provides an efficient and scalable alternative to existing approaches for time series counterfactual explanations.
\end{abstract}


\section{Introduction}
\label{sec:intro}

Machine-learning models are increasingly used to classify time series in
high-stakes domains such as healthcare monitoring, financial risk assessment,
and environmental forecasting
\cite{loh2022application,bharati2023review,vcernevivciene2024explainable}.
Although deep models achieve strong predictive performance, their black-box
nature limits transparency and trust \cite{dwivedi2023explainable}.
Counterfactual explanations (CFEs) address this limitation by identifying
plausible changes to an input that would produce a desired prediction
\cite{wachter2017counterfactual}. Such explanations show how the model's
decision responds to changes in the input rather than merely identifying
features associated with the original prediction.

Generating meaningful CFEs for time series is particularly challenging.
Unlike tabular features, time-series values form temporally dependent
trajectories whose trends, continuity, and recurring patterns must be
preserved. Pointwise or segment-level changes can therefore introduce abrupt
discontinuities or unrealistic shapes
\cite{delaney2021instance,li2022motif}. Optimization-based approaches seek to
improve validity and temporal coherence through combinations of prediction,
distance, and structural constraints
\cite{filali2022mining,li2022sg,hosseinzadeh2024acts,wang2024glacier}.
However, most existing approaches remain instance-based: they perform a new
iterative search for every test input, whether in the input space or a learned
latent space. Consequently, their computation grows with both the number of
instances and the cost of the optimization procedure, while knowledge gained
from one search is not reused for subsequent inputs. This makes it difficult
to jointly achieve validity, plausibility, and scalability.

\begin{figure*}[!t]
    \centering
    \includegraphics[width=0.8\linewidth]{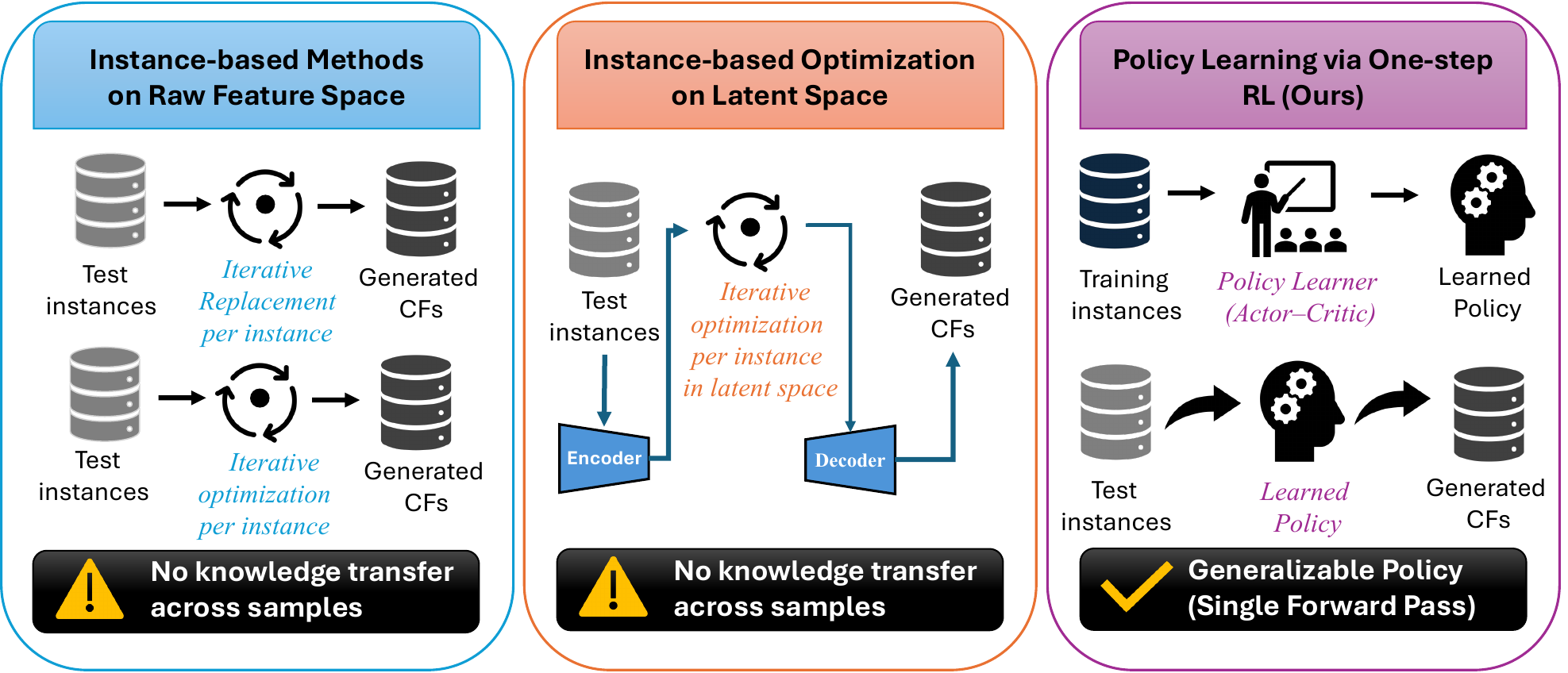}
    \caption{Comparison of instance-based counterfactual generation methods
    and the proposed policy-learning framework.}
    \label{fig:comparison}
\end{figure*}

To address this limitation, we formulate counterfactual generation as a
one-step reinforcement-learning problem, or contextual bandit. A pre-trained
autoencoder maps each input $x$ to a compact latent representation $z$. Given
$z$, the classifier's original prediction $y_M$, and a desired target class
$y_T$, an actor produces a counterfactual latent representation $z_{CF}$,
which the decoder maps back to the input space. The black-box classifier then
provides an immediate binary reward indicating whether the decoded candidate
is assigned to $y_T$. A critic learns a differentiable approximation of this
reward and guides the actor, while an $L_1+L_2$ proximity penalty limits
deviations from the original input. By learning from interactions collected
across many training instances, the actor acquires a reusable generation
policy instead of restarting an optimization procedure for each input. Once
trained, it generates counterfactuals in batches through a single actor
forward pass followed by decoding. The framework requires only classifier
predictions and thus does not require gradient access to the black-box model.

We evaluate the proposed framework on 13 datasets from the UCR Time Series
Classification Archive and compare it with optimization-based,
saliency-guided, and latent-space baselines. The results show high and stable
validity together with consistently strong plausibility under Isolation
Forest, Local Outlier Factor, and One-Class SVM scores. The method produces
larger aggregate $L_1$ and $L_2$ distances than localized input-space methods,
but remains competitive in $L_{\infty}$ and avoids extreme pointwise changes.
Runtime results further show that the benefit of the learned policy increases
with the number of test instances, making it particularly suitable for
large-scale counterfactual generation.

The main contributions of this work are as follows:
\begin{itemize}
    \item We formulate time-series counterfactual generation as a one-step
    contextual-bandit problem that learns a transferable policy and eliminates
    per-instance optimization at inference time.

    \item We develop a black-box actor--critic framework that generates
    counterfactuals in an autoencoder latent space and combines reward
    maximization with input-space proximity regularization.

    \item We evaluate validity, proximity, plausibility, ablation behavior,
    and runtime on 13 UCR datasets, demonstrating reliable generation and
    increasing scalability benefits as the test set grows.
\end{itemize}

\section{Related Work}
Counterfactual explanations (CFEs) have received increasing attention because of their ability to provide intuitive and actionable insights into machine learning predictions. By identifying minimal changes that alter a model’s decision, CFEs help users understand model behavior and explore alternative outcomes. Such explanations are especially valuable in high-stakes applications, where transparency and interpretability are as important as accuracy.

Early efforts in time series CFEs relied on instance-based heuristics, producing counterfactuals by perturbing each instance individually, often through segment replacement. For example, NG \cite{delaney2021instance} and MG-CF \cite{li2022motif} construct counterfactuals by altering or replacing salient subsequences drawn from the data (e.g., CAM-highlighted regions, class motifs). Their main strength is interpretability, since the modifications are grounded in real observed patterns that domain experts can readily relate to. In addition, because they reuse real subsequences, these methods often produce results that appear plausible at the local level. However, they depend heavily on the quality of saliency or motif mining and can struggle with validity (failing to consistently flip the label) and global coherence (as a swapped segment may not integrate smoothly with the rest of the sequence), which reduces their robustness across datasets.

Building on this idea, optimization-based approaches formulate counterfactual generation as an optimization problem for each input instance. The classic formulation by \cite{wachter2017counterfactual}, originally introduced for tabular data, has since been widely adopted as a baseline in time series studies \cite{li2022motif, li2022sg, bahri2022shapelet}. Wachter typically minimizes a loss that balances prediction validity (ensuring the label flips) with proximity to the original instance, often measured by L1 or L2 distance. While this formulation is simple and intuitive, proximity constraints alone cannot guarantee temporal plausibility, and naïve pointwise perturbations often disrupt sequential dynamics, leading to unrealistic counterfactuals. To address this limitation, constrained optimization variants have been proposed. For example, SG-CF \cite{li2022sg} and TeRCE \cite{bahri2022temporal} restrict perturbations to subsequences defined by shapelets or temporal rules, while TimeX \cite{filali2022mining} enforces temporal coherence through barycenter averaging. These constraints improve local interpretability and coherence but introduce additional rigidity, as they rely on predefined structures that may not capture global dependencies, and they still suffer from the inefficiency of per-instance optimization. 

A related direction uses saliency-guided masks to localize perturbations and improve interpretability. Methods such as CELS \cite{li2023cels} and Info-CELS \cite{li2025info} learn perturbation masks guided by gradient information and the nearest unlike neighbor, enabling sparse and interpretable modifications without the need for predefined structures or rules. These approaches produce intuitive visual explanations by highlighting learned saliency maps. However, a key limitation is their reliance on the nearest unlike neighbor: counterfactuals are generated by interpolating between the original instance and this neighbor, with the interpolation rate controlled by the saliency map. If the nearest unlike neighbor exhibits temporal shifts or misalignments, the resulting counterfactuals can still be implausible, even if the label flips successfully.

Latent-space perturbation has recently emerged as a promising direction for improving plausibility. Glacier \cite{wang2024glacier} perturbs representations in both the input space and autoencoder-derived latent space, guided by saliency explainers such as LimeSegment \cite{sivill2022limesegment}. The saliency vectors serve as constraints that determine which time steps should be perturbed, helping to localize modifications and preserve temporal structure. While this approach improves temporal localization, its performance remains highly sensitive to the quality of the saliency explainer. If the saliency vectors highlight noisy or irrelevant regions, the generated counterfactuals may still lack plausibility and robustness.

All of the methods discussed above are fundamentally instance-based: counterfactuals are generated independently for each test input via iterative replacement or optimization in either the raw feature space or a learned latent space. This design limits scalability and generalization, as the search must be repeated from scratch for every new instance. A small number of recent studies have explored reinforcement learning–based counterfactual generation in tabular domains, where agents learn action policies to apply minimal perturbations that flip model predictions \cite{samoilescu2021model, chen2022relax, nguyen2021counterfactual}. These works demonstrate the promise of experience-driven policy learning but rely on the assumption of independent, semantically meaningful features. Unlike tabular data, time series consist of temporally dependent trajectories, and treating each time step as an independent feature leads to temporally incoherent perturbations and an impractically large action space, resulting in unrealistic counterfactuals and poor scalability. Motivated by this gap, we revisit the RL perspective for time series classification. Instead of per-instance optimization or perturbing raw time steps, our approach learns a transferable policy in a structured latent space that captures temporal dependencies, enabling temporally coherent counterfactuals and scalable, single-pass inference on unseen inputs.






\section{Preliminary}
\subsection{Counterfactual Explanation for Time Series Classification}

We consider a dataset of $N$ time series
$\mathcal{X}=\{x_1,\ldots,x_N\}$, where each
$x_i\in\mathbb{R}^{T}$ is a sequence of length $T$. Let
$M:\mathbb{R}^{T}\rightarrow\mathcal{Y}$ be a black-box classifier
that maps each time series to one of $C$ class labels in
$\mathcal{Y}$. For an input instance $x$, the model prediction is
$y_M=M(x)$. Given a target label $y_T\neq y_M$, the goal of
counterfactual explanation is to find a perturbed time series
$x_{CF}$ such that
\begin{equation}
x_{CF}=x+\delta,\quad \delta\in\mathbb{R}^{T},\quad M(x_{CF})=y_T .
\end{equation}
The perturbation $\delta$ describes the changes needed to alter the
model prediction from $y_M$ to the desired target label $y_T$.




\section{Methodology}
\begin{figure*}[!t]
    \centering
    \includegraphics[width=0.8\linewidth]{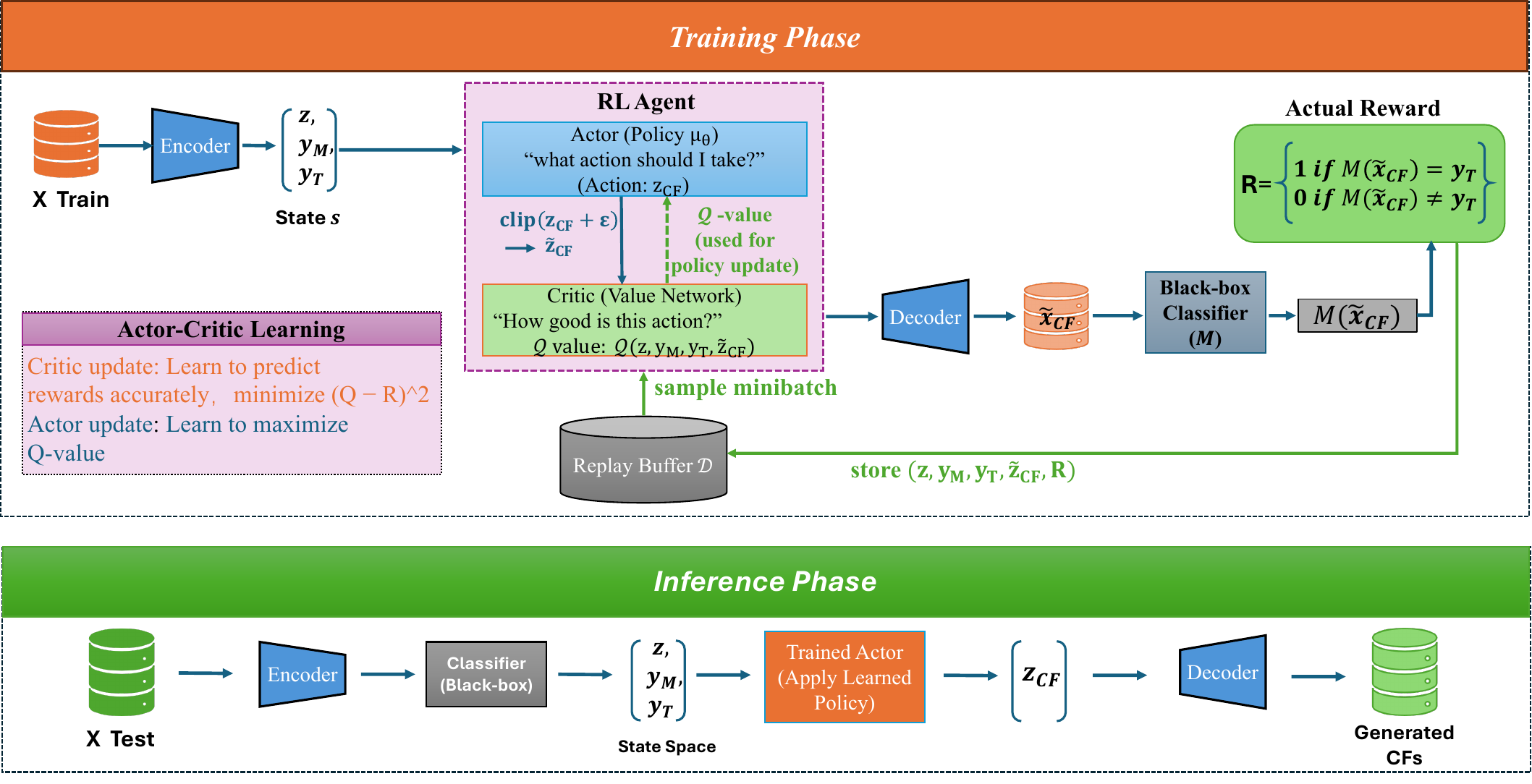}%
    \caption{
    The overview of the proposed framework.
    }
    \label{fig:rlcf_overview}
\end{figure*}
Instead of relying on instance-based optimization, where each counterfactual must be obtained by perturbing the input through the same optimization process for every instance, we employ an actor--critic reinforcement learning framework to learn a policy that generates counterfactuals efficiently. Once trained, the policy produces counterfactuals in batches through a single forward pass, offering greater scalability than per-instance optimization.

Formally, we formulate counterfactual generation as a \emph{one-step} Markov decision process (contextual bandit):
\begin{equation}
\mathcal{M} = (\mathcal{S}, \mathcal{A}, R), \quad s \in \mathcal{S}, \; a \in \mathcal{A}, \; R = R(s, a),
\end{equation}
where $\mathcal{S}$ is the state space, with $s = (z, y_M, y_T)$ representing the latent embedding $z$, current prediction $y_M$, and target class $y_T$; $\mathcal{A}$ is the action space, where $a = z_{CF}$ represents the modified latent embedding produced by the actor; and $R$ is the immediate reward function: $R(s,a) = 1$ if $M(\mathrm{dec}(a)) = y_T$, and $0$ otherwise.

Note that there is no transition function $P$ or next state $s'$: each counterfactual generation constitutes a single decision, after which the episode terminates. This differs from traditional multi-step RL, where agents navigate through sequential states. The environment in our formulation is the interaction between the time series representation and the classifier $M$. Given a state $s$, the actor generates an action $a$, which is decoded and evaluated by $M$ to produce an immediate binary reward. Unlike sequential RL problems with long-term credit assignment, our one-step formulation provides direct feedback after each decision.

To handle the high dimensionality and temporal structure of time series, we first train an autoencoder to obtain a lower-dimensional latent representation $z = \mathrm{enc}(x)$. Operating in this compact latent space significantly reduces the effective action dimensionality for the RL agent, avoiding the need to perturb every time step of the raw sequence. Counterfactual generation is then performed in this latent space, where perturbations are smoother and more structured. The actor network $\mu_\theta$ takes the latent state $z$ as input and outputs a perturbed embedding $z_{CF}$ corresponding to a counterfactual candidate. This representation is decoded back to the input space, $\tilde{x}_{CF} = \mathrm{dec}(z_{CF})$, which is then evaluated by the classifier to determine the reward.

We adopt a simplified deterministic actor--critic framework inspired by Deep Deterministic Policy Gradient (DDPG) \cite{lillicrap2015continuous}. The critic $Q_\phi(s,a)$ estimates the value of applying action $a$ in state $s$ and is trained by regressing directly to the observed (immediate) reward signal:
\begin{equation}
L_{\text{critic}} = \mathbb{E}\Big[\big(Q_\phi(s_t, a_t) - R_t\big)^2\Big].
\end{equation}

\textbf{Critic in One-Step Settings.} In standard multi-step DDPG, the critic estimates long-term cumulative rewards: $Q(s,a) = \mathbb{E}[R_t + \gamma R_{t+1} + \gamma^2 R_{t+2} + \cdots]$. However, in our one-step formulation, the critic simplifies to predicting immediate reward expectations:
\begin{equation}
Q(s,a) \approx \mathbb{E}[R \mid s, a] \quad \text{where } R \in \{0, 1\}.
\end{equation}
The critic loss becomes a direct regression to observed binary rewards without bootstrapping from future values. This differs from standard DDPG's temporal-difference target: $Q_{\text{target}} = R + \gamma Q'(s', a')$. We have no $s'$ (next state) or $\gamma$ (discount factor).

Despite this simplification, the critic plays three essential roles in our framework.
First, it reduces variance by learning a smooth function $Q(s,a)$ that approximates the expected reward, providing more stable gradients than the binary flip-reward signal.
Second, it enables off-policy learning through experience replay, allowing stored $(s,a,R)$ tuples to be reused efficiently even as the actor policy evolves.
Finally, it offers a differentiable surrogate for the non-differentiable classifier and reward, allowing the actor to be optimized via gradients of $Q(s,\mu_\theta(s))$.

The actor is trained to maximize the critic's estimate of value through the deterministic policy gradient:
\begin{equation}
L_{\text{max}} = -\, \mathbb{E}\big[Q_\phi(s_t, \mu_\theta(s_t))\big].
\end{equation}

To promote similarity between the counterfactual and the original input, we add a proximity regularizer that combines $L_1$ and $L_2$ distances:
\begin{equation}
\label{eq: prox}
L_{\text{prox}} = \beta \, \|\tilde{x}_{CF} - x\|_1 + \|\tilde{x}_{CF} - x\|_2^2.
\end{equation}

The final actor objective is then
\begin{equation}
L_{\text{actor}}^{\text{total}} = L_{\text{max}} + \lambda_{\text{p}}\, L_{\text{prox}}.
\end{equation}

During training, exploration is encouraged by injecting noise into the actor's proposed perturbed embedding, gradually transitioning from uniform random actions to noisy perturbations around the actor's output. Through this combination, the actor learns to generate counterfactuals that maximize validity while preserving proximity, and the decoder ensures that the resulting sequences remain realistic and temporally coherent.
\paragraph{Training Workflow.}
Algorithm~\ref{alg:training} and Figure \ref{fig:rlcf_overview} summarizes the training pipeline of our policy learning framework for counterfactual generation in time series, which is conducted on the training dataset. The procedure begins by loading a pre-trained encoder-decoder pair (line~1), which defines a structured latent space for applying perturbations. The actor and critic networks are initialized along with a replay buffer, and a reward function $f$ is defined to convert classifier predictions relative to the target label into a reward signal (lines~2--4).

\begin{algorithm}[t]
\caption{Training procedure}
\label{alg:training}
\textbf{Inputs:} $M$: black-box model; proximity loss weight $\lambda_p$; total steps $T$; 
update interval $U$; batch size $|B|$; exploration std.\ $\sigma$; mixing coefficient $\beta{=}0.5$\\
\textbf{Outputs:} Trained actor network $\mu$ for counterfactual generation\\
\textbf{Note:} Each training step generates \emph{one} counterfactual experience $(s, a, R)$ without sequential state transitions.
\begin{algorithmic}[1]
\State Load pre-trained encoder $enc$ and decoder $dec$
\State Initialize actor $\mu(\cdot;\theta_\mu)$ and critic $Q(\cdot;\theta_Q)$
\State Initialize replay buffer $\mathcal{D}$
\State Define reward function $f(\cdot,\cdot)$

\For{$t=1$ to $T$} \Comment{$T$ training steps; each generates ONE counterfactual experience}
  \State Sample input time series $x$ and target class $y_T$
  \State $y_M \gets M(x)$; \quad $z \gets enc(x)$
  \State $z_{CF} \gets \mu(z, y_M, y_T;\theta_\mu)$
  \State $\tilde z_{CF} \gets \text{clip}(z_{CF} + \varepsilon,\,-1,1)$, \; $\varepsilon \sim \mathcal{N}(0,0.1)$
  \State $\tilde x_{CF} \gets dec(\tilde z_{CF})$
  \State $R \gets f(M(\tilde x_{CF}), y_T)$ \Comment{Immediate binary reward: 1 if flip, 0 otherwise}
  \State Store $(z, y_M, y_T, \tilde z_{CF}, R)$ in the replay buffer $\mathcal{D}$ \Comment{No next state---episode ends}

  \If{$t \bmod U = 0$}

      \State Sample batch $\mathcal{B}$ of size $B$ from $\mathcal{D}$
      \State Update critic by one-step gradient descent using:
      \[
      \nabla_{\theta_Q}\; \frac{1}{|B|} \!\sum_{B}
      \big(Q(z, y_M, y_T, \tilde z_{CF}) - R\big)^2
      \]
      \State Recompute $z_{CF} \gets \mu(z, y_M, y_T;\theta_\mu)$,\; $x_{CF} \gets dec(z_{CF})$

      \[
      L_{\max} = -\tfrac{1}{|B|}\sum_{B} Q(z, y_M, y_T, z_{CF})
      \]
      \[
      L_{\text{prox}} = \tfrac{1}{|B|}\sum_{B} \big(\beta\,\|x - x_{CF}\|_1 + \,\|x - x_{CF}\|_2^2\big)
      \quad \text{with } \beta=0.5
      \]
      \State Update actor by one-step gradient descent using:
      \[
      \nabla_{\theta_\mu} \big(L_{\max} + \lambda_p\, L_{\text{prox}}\big)
      \]

  \EndIf
\EndFor

\end{algorithmic}
\end{algorithm}



At each training step (lines~5--12), an input time series $x$ and target label $y_T$ are sampled. The input is encoded into a latent representation $z$, and the actor generates a candidate counterfactual latent $z_{CF}$ conditioned on $(z, y_M, y_T)$. Gaussian noise is added for exploration, and the perturbed latent $\tilde z_{CF}$ is clipped before being decoded into a candidate $\tilde x_{CF}$. The black-box classifier $M$ evaluates $\tilde x_{CF}$ to produce a reward $R = f(M(\tilde x_{CF}), y_T)$, and the interaction tuple $(z, y_M, y_T, \tilde z_{CF}, R)$ is stored in the replay buffer.

Every $U$ steps, the framework performs a gradient update (lines~13–19). First, a minibatch from the replay buffer is used to update the critic by minimizing the squared error between its predicted value and the observed reward, thereby improving its approximation of the reward function. Next, the actor is updated with a combined objective: a policy loss, which encourages the actor to maximize the critic’s Q-value and thus generate valid counterfactuals, and a proximity loss, which penalizes large deviations from the original input. The overall actor loss $L_{\text{max}} + \lambda_p L_{\text{prox}}$ balances validity with proximity.

Through repeated interactions with the black-box classifier and joint optimization of both actor and critic, the actor progressively learns a policy for generating counterfactuals that achieve the desired label change while remaining close to the original time series.

\paragraph{Inference Workflow.}
As illustrated in Figure~ \ref{fig:rlcf_overview}, after training, counterfactual generation is performed by applying the learned policy directly to each test instance. Given an input time series $x$, the encoder first maps it to a latent representation $z=\mathrm{enc}(x)$, and the black-box classifier provides the original prediction $y_M=M(x)$. Conditioned on $(z,y_M,y_T)$, where $y_T$ is the desired target label, the trained actor outputs a counterfactual latent representation $z_{CF}=\mu(z,y_M,y_T)$. The decoder then maps this representation back to the input space, producing the counterfactual sequence $x_{CF}=\mathrm{dec}(z_{CF})$. Since this process requires only one actor forward pass followed by decoding, inference avoids per-instance optimization and supports efficient batch generation of counterfactual explanations.

\section{Experiments}
\subsection{Experimental Setup}
\begin{table}
\centering
\caption{Datasets Metadata}
\resizebox{\linewidth}{!}{
\begin{tabular}{|c||c|c|c|c|c|}
\hline
\label{tab:UCR}
ID & Dataset Name  & $\mathcal{L}$ & DS train size & DS test size & Type \\ 
\hline
0  & Chinatown                   & 24   & 20    & 343  & TRAFFIC \\ 
\hline
1  & Coffee                      & 286  & 28    & 28   & SPECTRO \\ 
\hline
2  & ECG200                      & 96   & 100   & 100  & ECG \\ 
\hline
3  & FordA                       & 500  & 3601  & 1320 & SENSOR \\ 
\hline
4  & FordB                       & 500  & 3636  & 810  & SENSOR \\ 
\hline
5  & FreezerRegularTrain         & 301  & 150   & 2850 & DEVICE \\ 
\hline
6  & GunPoint                    & 150  & 50    & 150  & HAR \\ 
\hline
7  & GunPointAgeSpan             & 150  & 135   & 316  & HAR  \\ 
\hline
8  & GunPointMaleVersusFemale    & 150  & 135   & 316  & HAR  \\ 
\hline
9  & GunPointOldVersusYoung      & 150  & 135   & 316  & HAR   \\ 
\hline
10 & HandOutlines                & 2709 & 1000  & 370  & IMAGE  \\ 
\hline
11 & TwoLeadECG                  & 82   & 23    & 1139 & ECG    \\ 
\hline
12 & Wafer                       & 152  & 1000  & 2000 & SENSOR  \\ 
\hline
\end{tabular}
}
\footnotesize { $\mathcal{L}$: sequence length.}
\end{table}
\textbf{Datasets. }We evaluate our approach on a diverse collection of publicly available univariate time series datasets drawn from the University of California, Riverside (UCR) Time Series Classification Archive \cite{dau2019ucr}. The UCR archive is a widely used benchmark suite for time series classification and has been extensively adopted to assess both traditional and deep learning models. To ensure that our findings are representative across different domains and levels of complexity, we select 13 datasets spanning application areas such as spectro, ECG, human activity recognition (HAR), sensor, traffic, etc. Table~\ref{tab:UCR} provides detailed statistics for all datasets, including sequence length, number of classes, and train–test splits. These datasets have been shown to yield strong performance with state-of-the-art deep classifiers \cite{ismail2019deep}, making them suitable for evaluating the quality of counterfactual explanations.  

The chosen datasets also vary in size, covering scenarios from small-scale problems with fewer than 100 training instances to medium-scale collections with 100–250 training samples, and to larger-scale datasets containing thousands of examples. This diversity allows us to systematically investigate how our method scales with training set size. Importantly, the datasets are partitioned following the standard UCR setup, where training and testing sets are evenly sampled across classes to avoid class imbalance issues.

\textbf{Baselines.}
We select four representative instance-based methods covering the main counterfactual-generation strategies.

\textit{Wachter} \cite{wachter2017counterfactual} directly optimizes counterfactuals in the raw input space using validity and proximity losses, providing a basic comparison with optimization without policy learning.

\textit{TimeX} \cite{filali2022mining} adds Dynamic Barycenter Averaging to per-instance optimization, representing methods with explicit temporal-plausibility constraints.

\textit{Info-CELS} \cite{li2025info} learns a continuous saliency map to interpolate between an input and its nearest unlike neighbor, representing sparse and localized perturbations.

\textit{Glacier-AE} \cite{wang2024glacier} performs per-instance counterfactual optimization in an autoencoder latent space without learning a generation policy, allowing comparison with our latent-space policy-learning framework.

\textbf{Evaluation Metrics.} Following the evaluation practices commonly adopted in the time-series counterfactual explanation literature \cite{li2026counterfactual}, we assess different methods along three standard metrics: 
\textit{(1) Validity Metric (Flip Label Rate)}: This metric measures how often the generated counterfactuals successfully change the model's original prediction. Formally, it is defined as:
\begin{equation}\label{equ:flipr}
    flip\_rate = \frac{num\_flipped}{num\_testsample},
\end{equation}
where \textit{num\_flipped} is the number of counterfactuals that flip the predicted label, and \textit{num\_testsample} is the total number of test inputs. A higher flip rate (closer to 1) indicates better validity, as more counterfactuals satisfy the label change requirement.

\textit{(2) Proximity Metrics ($L_1$, $L_2$, and $L_{\infty}$). }Proximity evaluates how close each counterfactual is to the original input. We report three common distance-based metrics: (a) the $L_1$ Distance (Manhattan distance) measures the total absolute change across all time steps; (b) the $L_2$ Distance (Euclidean distance): Emphasizes larger differences by squaring deviations; (c) the $L_{\infty}$ Distance (Chebyshev distance): Captures the maximum absolute deviation.





\textit{(2) Plausibility Metrics. (IF, LOF, OC\_SVM)}
We evaluate plausibility by comparing the generated counterfactuals against the training dataset and computing how ``out-of-distribution'' they appear under three standard unsupervised anomaly detectors: (a) \textbf{Isolation Forest (IF)} \cite{liu2008isolation}; (b) \textbf{Local Outlier Factor (LOF)} \cite{breunig2000lof}; (c) \textbf{One-Class SVM (OC\_SVM)} \cite{scholkopf2001estimating}. Each detector is trained only on the original training data, which defines the normal data manifold. We then feed the generated counterfactuals to these detectors and measure their outlier scores. Counterfactuals that receive lower outlier scores are considered more plausible, since they lie closer to the distribution of real time series in the training set.

\subsection{Implementation Details}
We train the framework for 50{,}000 steps using dataset-specific batch sizes ranging from 16 to 256 and set $\lambda_p=1$. For each dataset, we evaluate autoencoders with latent dimensions of $\{8,16,32,64\}$ and select the model with the lowest reconstruction error. The actor and critic are implemented as multilayer perceptrons with two 256-unit hidden layers, layer normalization, and ReLU activations. The actor uses a $\tanh$ output layer to generate bounded latent representations, while the critic outputs a scalar value. The critic minimizes the squared error between predicted values and observed rewards, whereas the actor maximizes the critic output with proximity regularization. Gaussian noise is added to the actor output for exploration. The black-box classifier and pre-trained autoencoder remain fixed during policy training. Training metrics and intermediate samples are logged using Weights \& Biases every 100 steps. The source code provides the complete architecture, dataset-specific parameters, and training configurations.\footnote{\url{https://github.com/Luckilyeee/PLCF}}

\subsection{Experimental Results}
\begin{figure}[!htbp]
    \centering
    \includegraphics[width=0.75\linewidth]{  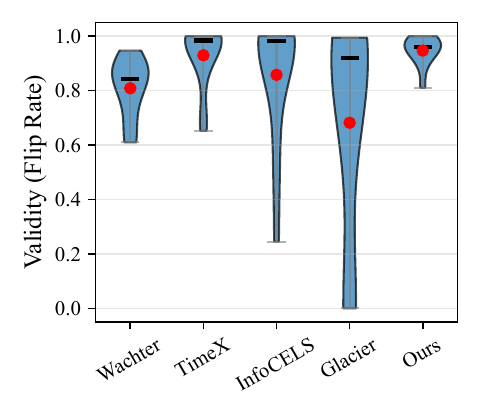}
    \caption{Validity evaluation measured by flip rate, where higher values indicate more valid counterfactuals.}
    \label{fig:validity}
\end{figure}

\begin{figure*}[!htbp]
    \centering
    \includegraphics[width=0.9\linewidth]{  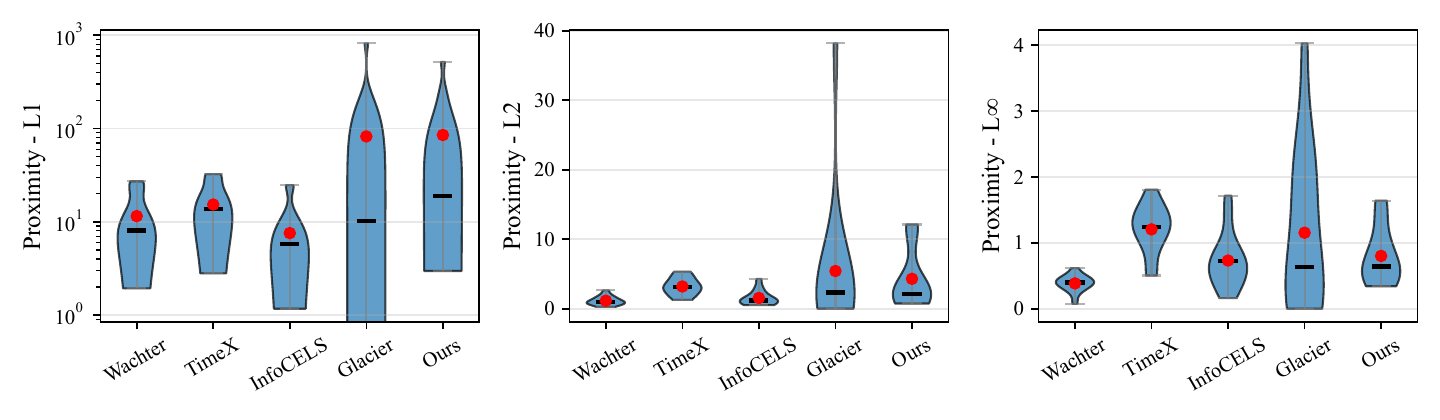}
    \caption{Proximity evaluation using three distance metrics. Lower values indicate counterfactuals closer to the original inputs.}
    \label{fig:proximity}
\end{figure*}

\begin{figure*}[tbp]
    \centering
    \includegraphics[width=0.9\linewidth]{  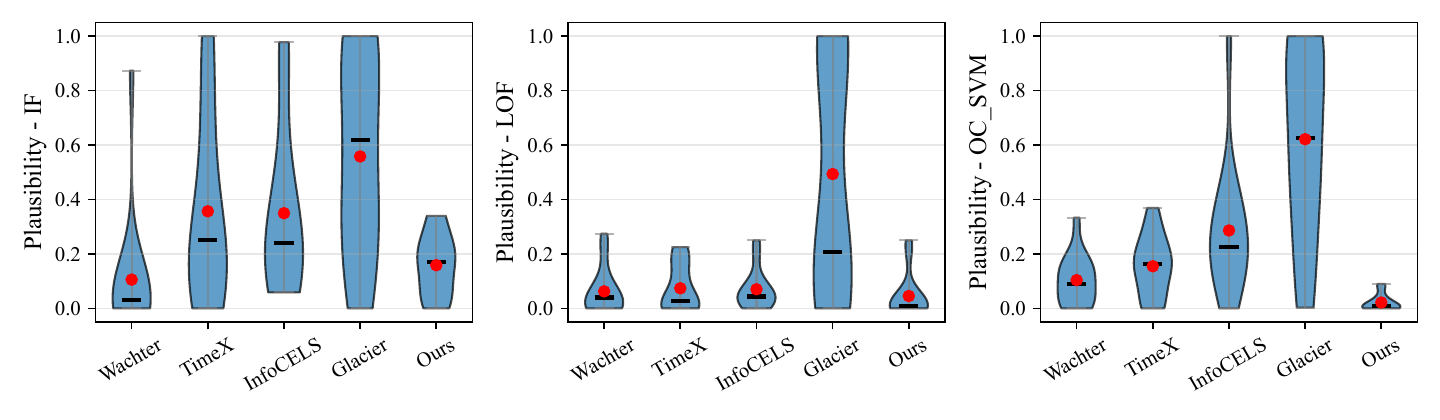}
    \caption{Plausibility evaluation using three outlier detection methods. Lower scores indicate more realistic counterfactuals.}
    \label{fig:plausibility}
\end{figure*}

\textbf{Validity (Flip Rate): }Figure~\ref{fig:validity} compares the validity of counterfactuals across methods, measured by flip rate. The violin plot shows the distribution of flip rates across datasets. (\textit{The black horizontal line and red dot represent the median and mean across datasets, respectively. This is kept consistent across all violin plots in the paper}.) Our policy learning framework consistently achieves strong performance, with validity concentrated in the 0.8–1.0 range and a high mean and median (~0.95), reflecting reliable performance across datasets. This indicates that nearly all generated counterfactuals successfully flip the classifier’s decision, a critical property for actionable explanations. TimeX and InfoCELS often reach near-perfect validity ($\approx$1.0), but their performance is less reliable, dropping below 0.7 on certain datasets. Glacier is even more fragile: while it can achieve high validity on some datasets, it collapses to zero on others. These fluctuations reduce the overall robustness of the baselines. By contrast, our proposed method maintains a compact distribution, whereas the wider spread of InfoCELS, TimeX, and Glacier reveals susceptibility to dataset-specific failures. Taken together, these results highlight that our proposed method provides a stable and generalizable mechanism for generating valid counterfactuals in diverse time series domains.

\textbf{Proximity (L1, L2, $L_{\infty}$):}
Figure~\ref{fig:proximity} reports the proximity of generated counterfactuals, where lower values indicate smaller perturbations from the original inputs. For L1 and L2 distances, InfoCELS, TimeX, and Wachter achieve the lowest values. InfoCELS attains the best proximity by perturbing only the most salient time steps, while TimeX and Wachter maintain relatively low distances through explicit distance-based penalties in their optimization objectives. By contrast, our method and Glacier operate in latent space rather than directly perturbing the input, which leads to higher L1 and L2 values after decoding, as modifications are distributed more broadly across the sequence. Although our method employs an elastic distance penalty within the actor network to constrain perturbations, the decoding process through the autoencoder inevitably spreads changes across multiple time steps, yielding larger aggregate distances.

For $L_{\infty}$, which captures the largest pointwise deviation, our method ranks behind Wachter but demonstrates competitive performance with TimeX and InfoCELS, while clearly outperforming Glacier. This suggests that although our method produces broader perturbations—reflected in its higher $L_{1}$ and $L_{2}$ distances—it effectively avoids extreme spikes at individual time steps, leading to smoother and more controlled modifications. This behavior stems from the method’s design: the autoencoder embedding promotes globally distributed adjustments rather than highly localized changes. As a result, our method yields counterfactuals that are less proximate in aggregate distance but remain better aligned with the underlying data manifold.

\begin{figure*}[!tbp]
    \centering

    \begin{subfigure}{0.48\textwidth}
        \centering
        \includegraphics[width=\linewidth]{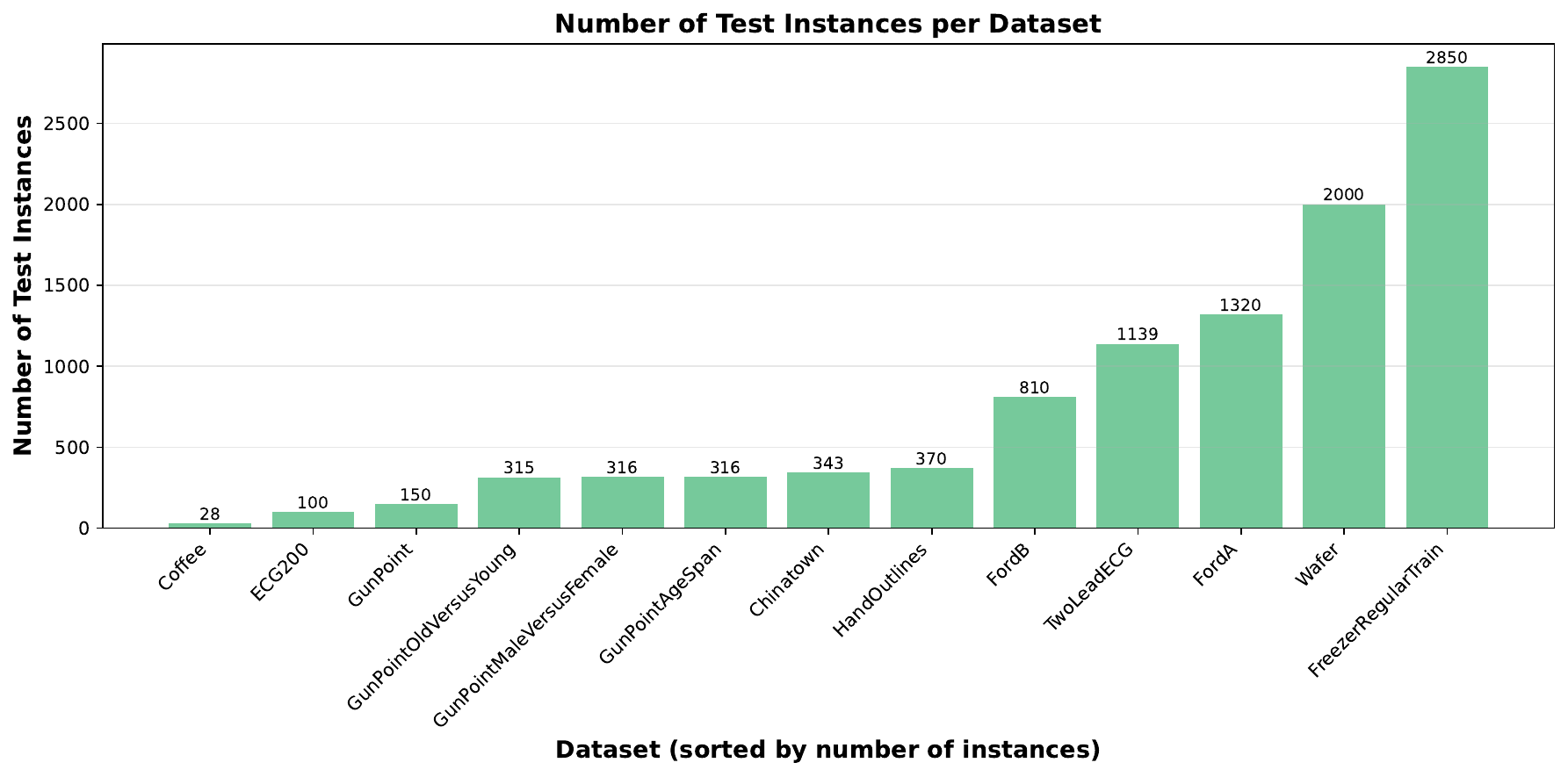}
        \caption{Dataset sizes across all UCR datasets}
        \label{fig:stats}
    \end{subfigure}
    \hfill
    \begin{subfigure}{0.48\textwidth}
        \centering
        \includegraphics[width=\linewidth]{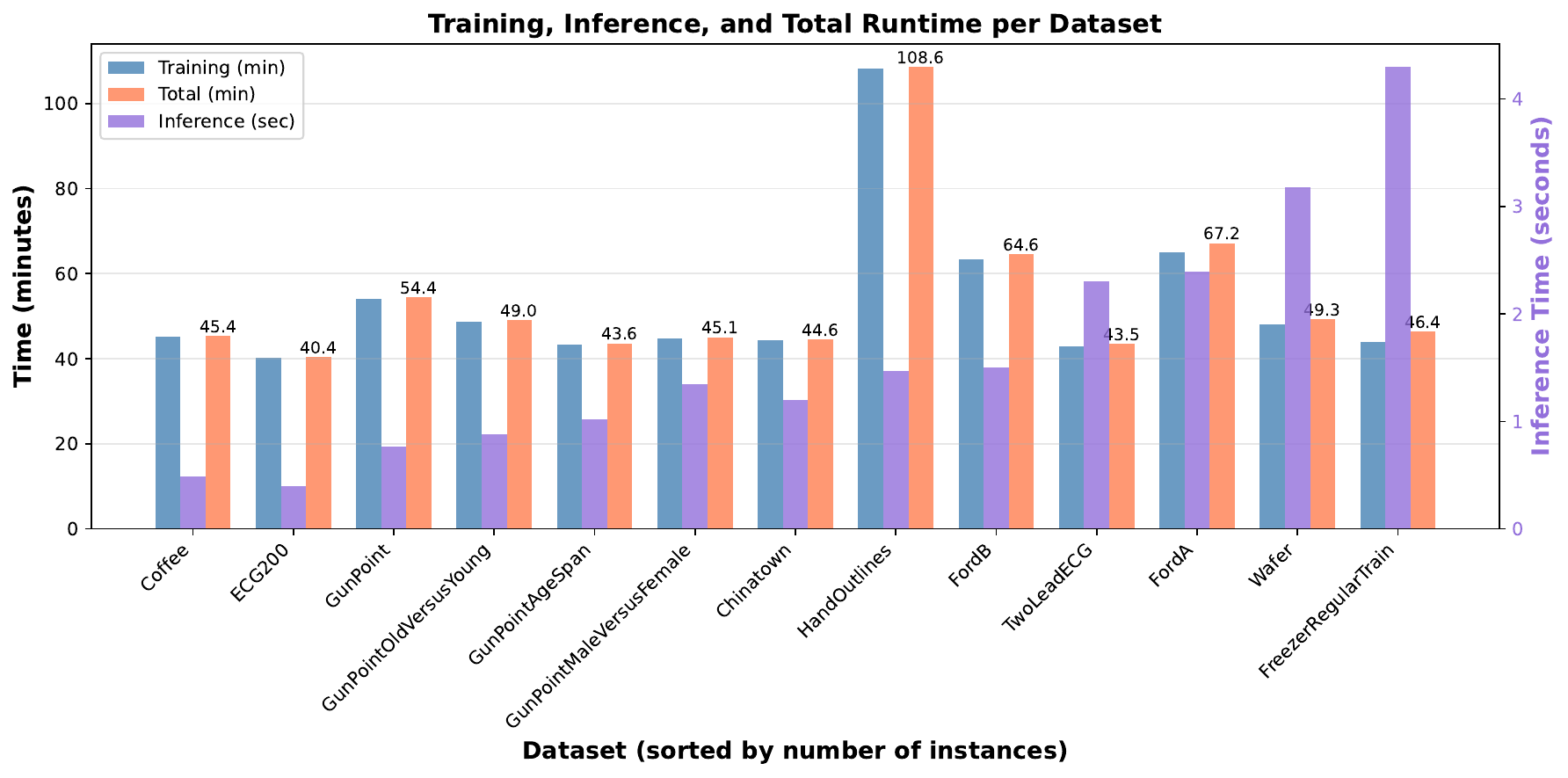}
        \caption{RL-CF (Ours) running time}
        \label{fig:rlcf}
    \end{subfigure}

    \begin{subfigure}{0.48\textwidth}
        \centering
        \includegraphics[width=\linewidth]{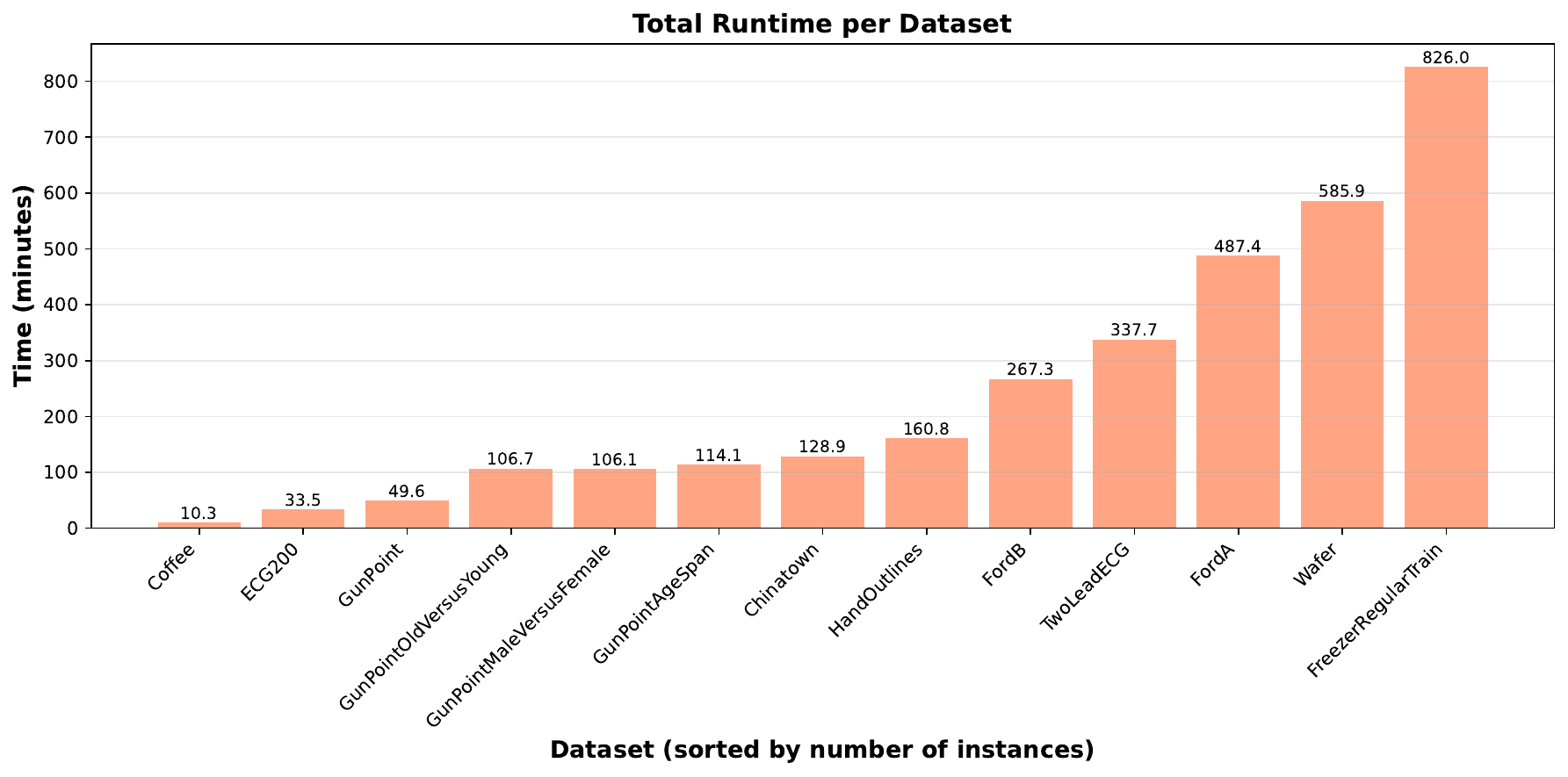}
        \caption{InfoCELS running time}
        \label{fig:info}
    \end{subfigure}
    \hfill
    \begin{subfigure}{0.48\textwidth}
        \centering
        \includegraphics[width=\linewidth]{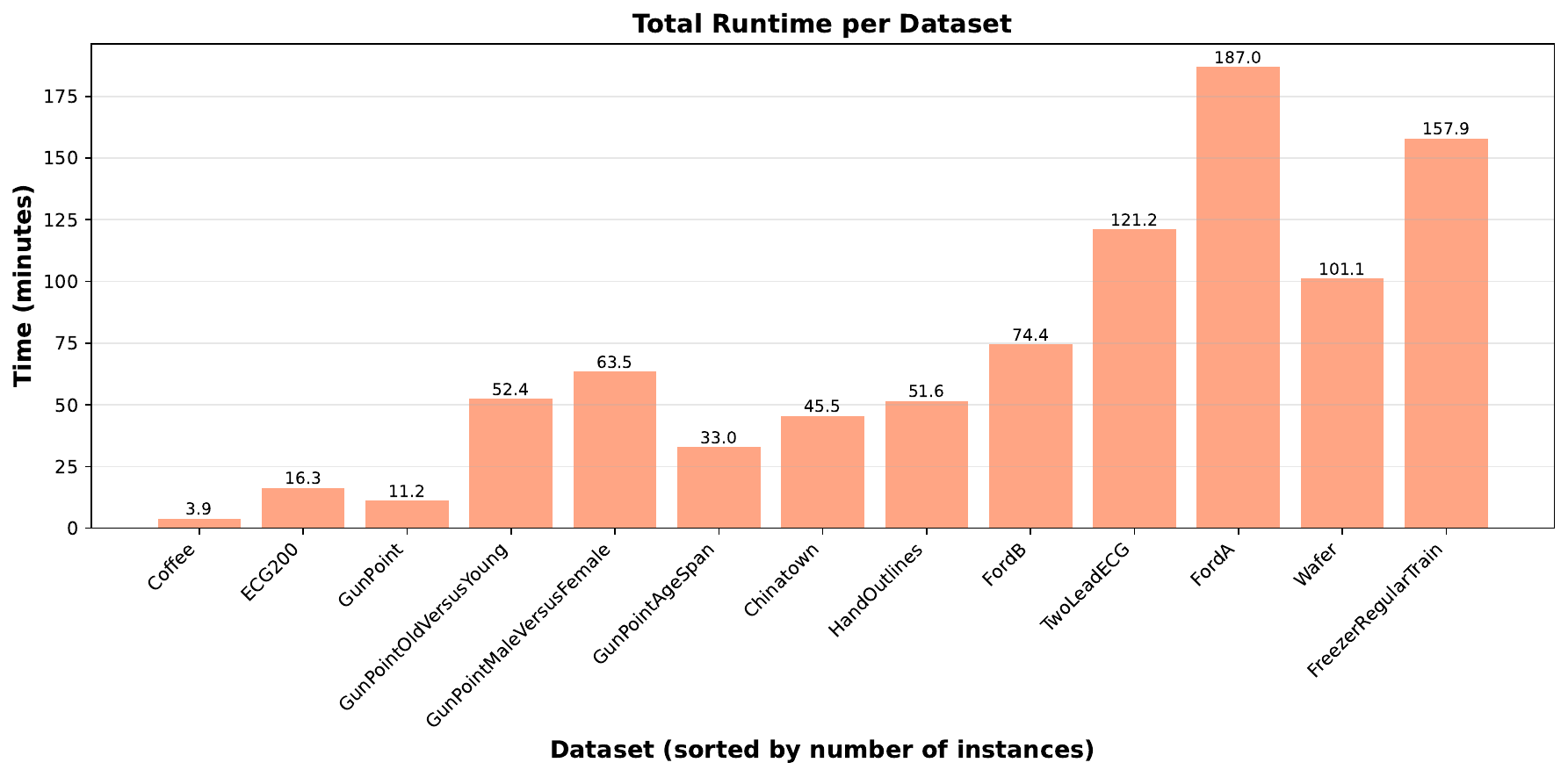}
        \caption{Glacier running time}
        \label{fig:glacier}
    \end{subfigure}


    \begin{subfigure}{0.48\textwidth}
        \centering
        \includegraphics[width=\linewidth]{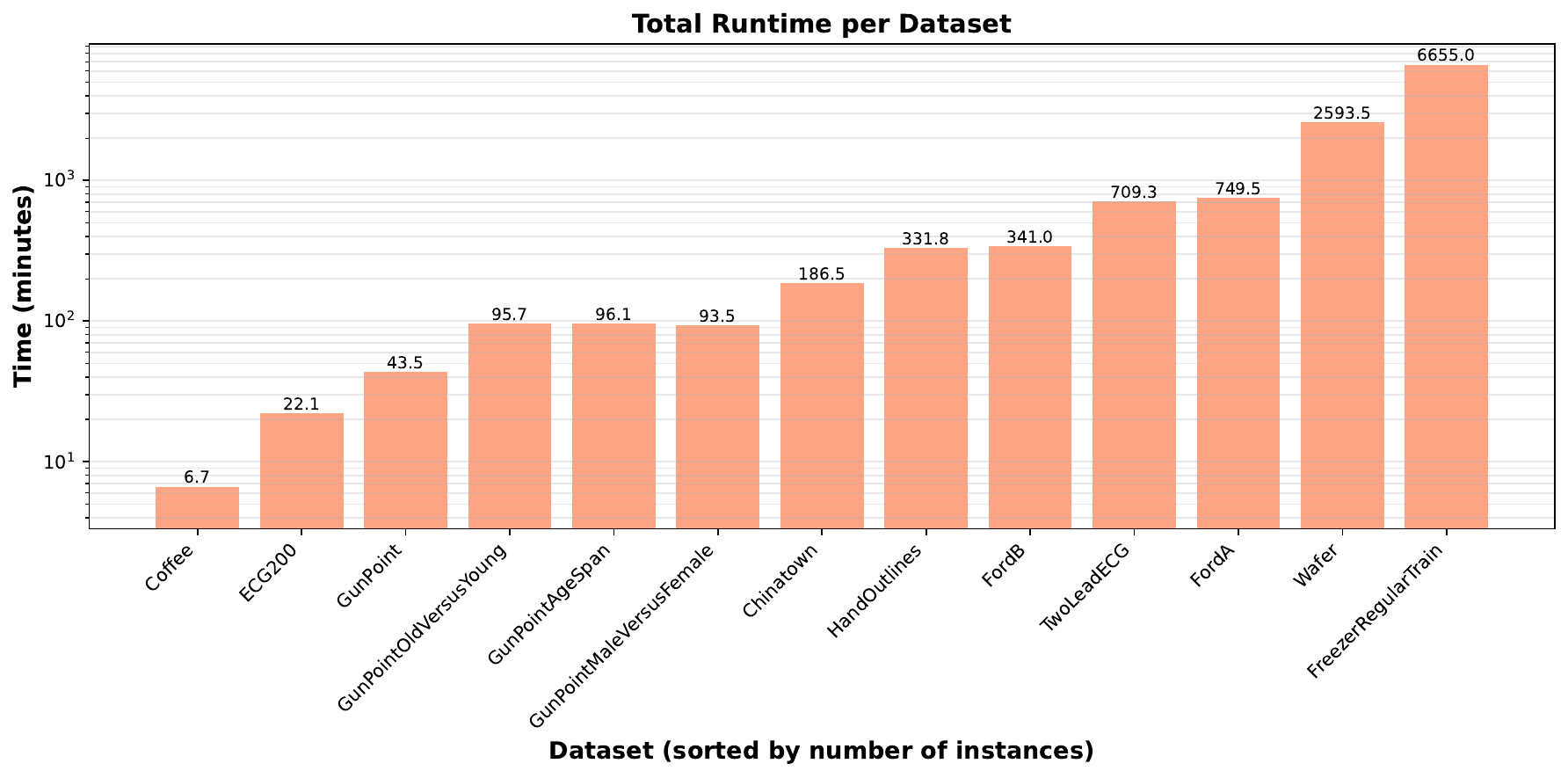}
        \caption{Wachter running time}
        \label{fig:alibi}
    \end{subfigure}
    \hfill
    \begin{subfigure}{0.48\textwidth}
        \centering
        \includegraphics[width=\linewidth]{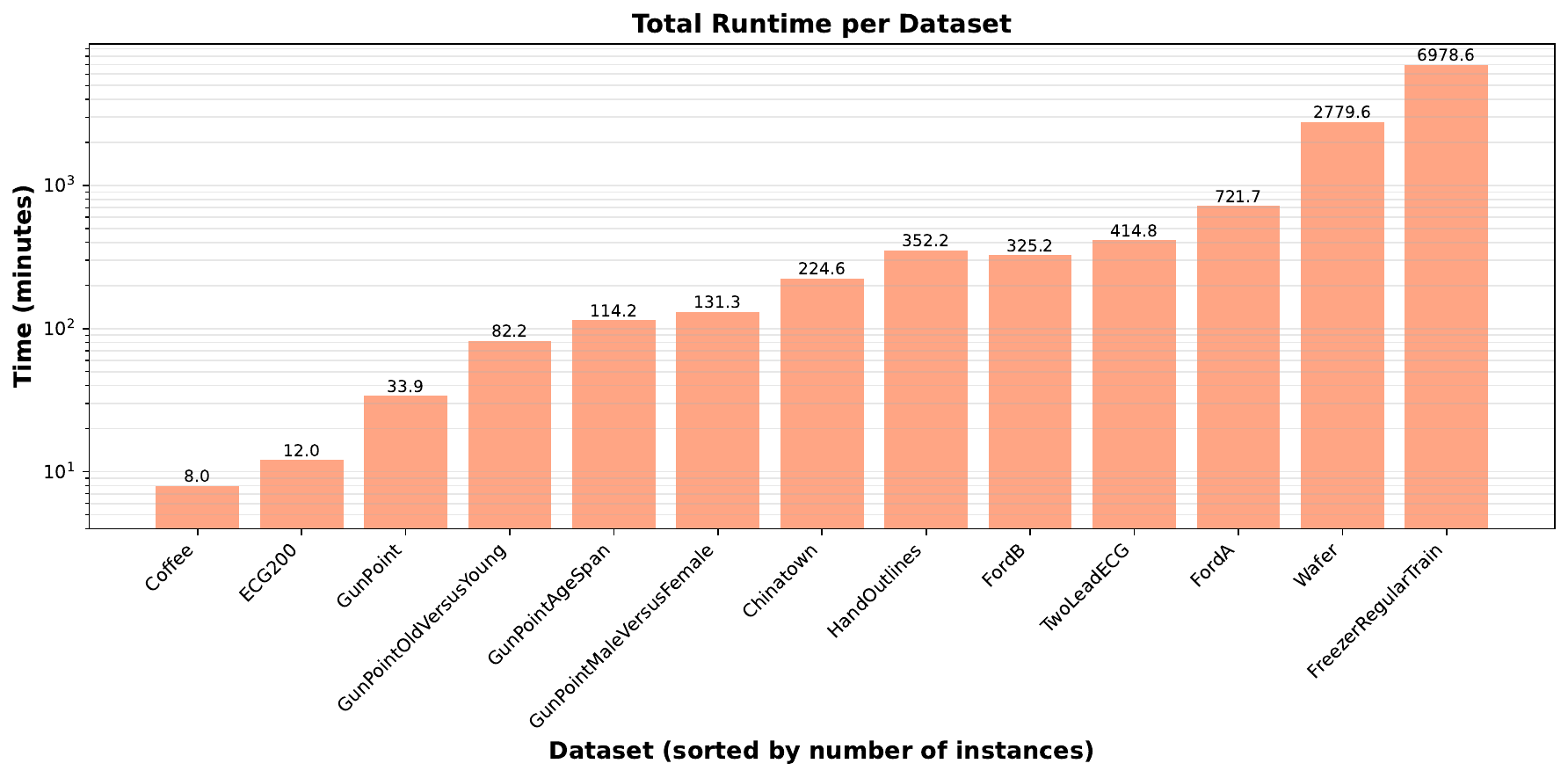}
        \caption{TimeX running time}
        \label{fig:timex}
    \end{subfigure}

    \caption{
        Running-time comparison of our proposed method and baseline counterfactual explanation methods across all UCR datasets.
        Subfigure~(a) shows the dataset sizes.
        Subfigure~(b) presents the running time of our proposed method.
        Subfigures~(c)--(f) report the runtime of InfoCELS, Glacier, Wachter, and TimeX, respectively.
        All results were obtained using the same hardware environment.
    }
    \label{fig:runtime_comparison}
\end{figure*}

\textbf{Plausibility (IF, LOF, OC\_SVM): }
Figure~\ref{fig:plausibility} shows that our method maintains consistently low outlier scores across all three detectors. Under IF, RL remains within the $0$–$0.35$ range, with both mean and median around $0.15$, while the baselines span the full $0$–$1$ interval with higher variance and frequent outliers. For LOF, our method stays below $0.3$ across all datasets, achieving the lowest mean and median values (close to $0$) and performing competitively with Wachter, TimeX, and InfoCELS, whereas Glacier again reaches up to $1.0$. Under OC\_SVM, our method clearly outperforms all baselines, keeping scores consistently below $0.1$ with mean and median near zero, while Wachter and TimeX extend toward $0.4$ and Glacier fluctuates widely between $0$–$1$.

These results demonstrate that the counterfactuals generated by our method align more closely with the data manifold, benefiting from latent-space policy learning that encourages globally coherent perturbations rather than localized and potentially unrealistic changes. By operating in the latent space, our method captures structural regularities of the data, which translates into counterfactuals that are not only valid but also plausible under multiple outlier-detection metrics. 

Taken together with the proximity results in Figure~\ref{fig:proximity}, these findings reveal a clear validity--proximity--plausibility trade-off. Input-space baselines such as Wachter, TimeX, and InfoCELS can achieve smaller aggregate $L_1$ and $L_2$ distances by making sparse or localized modifications, but these lower distances do not necessarily imply higher realism, as the resulting counterfactuals may still deviate from the data distribution. In contrast, our latent-space policy tends to produce broader perturbations after decoding, which can increase aggregate norm-based distances but helps preserve global temporal coherence and manifold consistency. Overall, the evidence highlights our proposed policy learning framework as a method that balances validity, proximity, and plausibility: it may sacrifice some raw input-space proximity, but it produces counterfactuals that are more stable, distributionally plausible, and effective for model explanation.

\begin{figure*}[!tbp]
    \centering
    \includegraphics[width=0.8\linewidth]{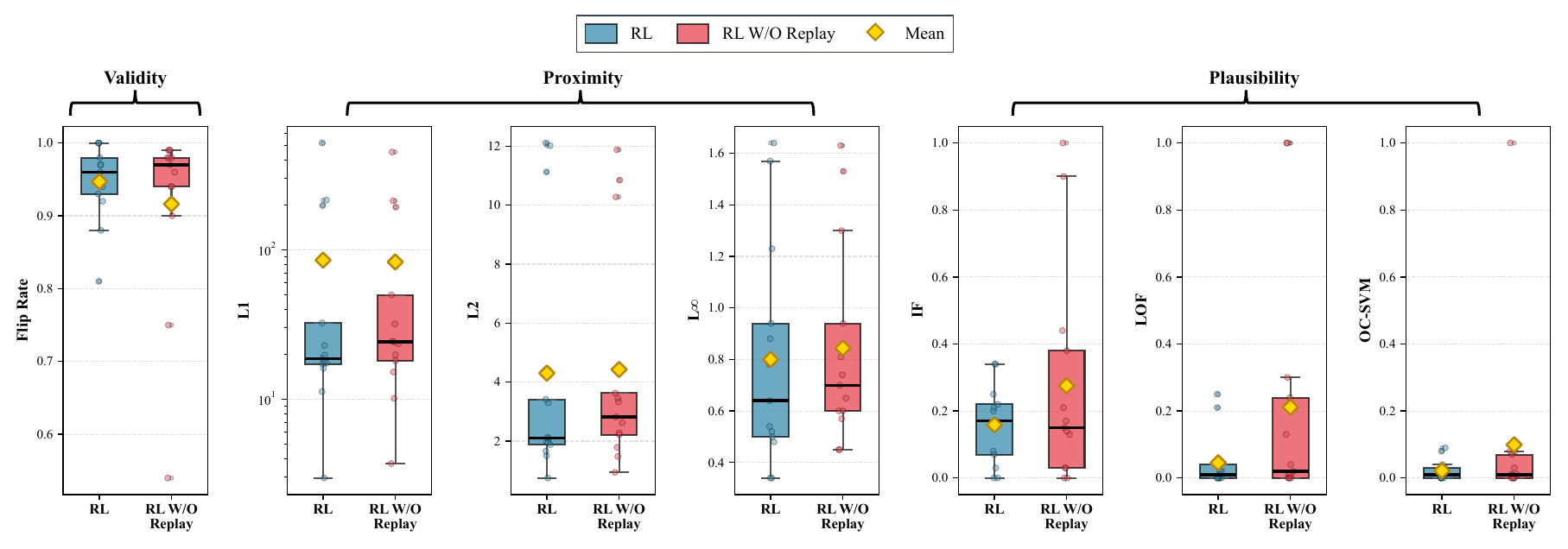}
    \caption{Ablation study comparing the full RL method with experience replay against the variant without replay.}
    \label{fig: ablation study}
\end{figure*}

\subsection{Efficiency Analysis and Running-Time Comparison}
\label{appendix:runtime}
To compare the computational efficiency of different counterfactual generation approaches, 
we analyze the runtime of our proposed framework relative to mainstream instance-based 
optimization baseline methods. A key distinction is that baseline methods do not include a separate 
``training'' phase for counterfactual generation---they optimize each test instance 
independently---whereas our proposed approach consists of both a training stage (policy learning) 
and a fast inference stage (single forward pass). To provide a fair comparison, we report:
\begin{itemize}
    \item \textbf{Training time} (RL only),
    \item \textbf{Inference time} (all methods), and
    \item \textbf{Total end-to-end time} (RL training + RL inference vs.\ inference-only baselines).
\end{itemize}

Figure~\ref{fig:runtime_comparison} presents the runtime comparison across datasets of increasing size 
(ordered from smallest to largest). Sorting datasets by size makes it easier to observe how 
methods scale as the number of test samples grows. Because all instance-based methods require 
iterative optimization for every input, their runtime increases roughly linearly with dataset 
size. In contrast, the RL method is trained once on the dataset, and its training cost is independent of the number of test instances. During inference, each counterfactual is produced by a single actor forward pass, which takes only milliseconds, so the overall inference time grows linearly but very slowly with the size of the test set.

\paragraph{Small Datasets.}
On small datasets (e.g., \texttt{Coffee}, \texttt{ECG200},\texttt{GunPoint}), instance-based methods such as Wachter, TimeX, InfoCELS, and Glacier have low total runtime simply because there are very few test samples. In these cases, the one-time training cost of our method becomes the main part of the total runtime, so the baselines may look faster. This is expected, since the benefit of training an RL policy only shows up when the test set is larger.

\paragraph{Medium-Sized Datasets.}
As we move to medium-sized datasets, the difference becomes more pronounced. Instance-based methods require optimization loops for an increasing number of samples, which causes their inference time to grow rapidly. In contrast, our method is trained once and its training cost stays within a reasonable range for all datasets. During inference, our method generates counterfactuals with a single forward pass, so inference time only grows linearly in the number of test instances and stays within seconds. As a result, the total runtime of our method becomes competitive with—and often lower than—the baseline methods on medium-sized datasets.

\paragraph{Large Datasets.}
For the largest datasets (e.g., \texttt{FordA}, \texttt{Wafer}, \texttt{FreezerRegularTrain}), the scalability benefits of 
Our method becomes most significant. Baseline methods must run thousands of optimization loops, causing their runtime to increase sharply. Wachter, TimeX, InfoCELS, and Glacier all show substantial slowdowns because each counterfactual requires repeated gradient-based updates on every instance. In contrast, our method introduces no per-instance optimization: once trained, generating a counterfactual only requires an actor forward pass followed by decoding. Although RL inference still grows linearly with the number of test samples, each forward pass takes only milliseconds, making the overall inference cost very small. As a result, our method achieves the lowest total runtime among all approaches on large datasets. These results highlight that our method is especially advantageous for large-scale counterfactual generation scenarios.

\subsection{Ablation Study}

To evaluate the contribution of \textit{experience replay} in our counterfactual generation framework, we conduct an ablation study comparing the full method (RL agent with the replay buffer) with a variant that removes the replay buffer (RL w/o Replay). The goal of this ablation is to examine how replay affects the performance of generated counterfactuals across three key metrics: validity, proximity, and plausibility. 

From Figure~\ref{fig: ablation study}, we observe that experience replay contributes differently across the three evaluation metrics, with varying levels of impact on validity, proximity, and plausibility. Importantly, both proximity and plausibility are computed \textit{only on valid counterfactuals}, ensuring that these results directly reflect the quality of CFs that successfully flip the label.

For validity, replay helps produce a tighter and higher flip-rate range of $[0.8, 1.0]$, whereas the no-replay variant has a wider range of $[0.5, 0.99]$. Although the median validity without replay is slightly higher and the mean slightly lower than with replay, this indicates that not all datasets require replay to maintain high validity of counterfactuals generated; some datasets could generate high validity counterfactuals even without it.  For proximity, all three metrics ($L_1$, $L_2$, and $L_\infty$) show only minor differences between the replay and no-replay variants. The values are generally comparable across both settings, with no large or systematic shift. The replay variant exhibits slightly lower mean and median values across datasets, but the overall effect is modest. This indicates that replay does not strongly influence how close the counterfactuals remain to the original inputs. The largest performance difference appears in the plausibility metrics (IF, LOF, and OC-SVM). Even though the no-replay variant produces \textit{valid} counterfactuals (i.e., they successfully flip the label), these valid CFs exhibit noticeably higher OOD scores and substantially larger variance, indicating that the perturbations are more likely to deviate from the data manifold. In comparison, the replay variant consistently produces lower and more stable plausibility values. These results collectively show that experience replay plays an important role in generating valid counterfactuals that are also plausible and manifold-consistent.

\subsection{Case Study: Qualitative Comparison of Counterfactual Explanations}
\begin{figure}[tbp]
        \centering
        \includegraphics[width=0.9\linewidth]{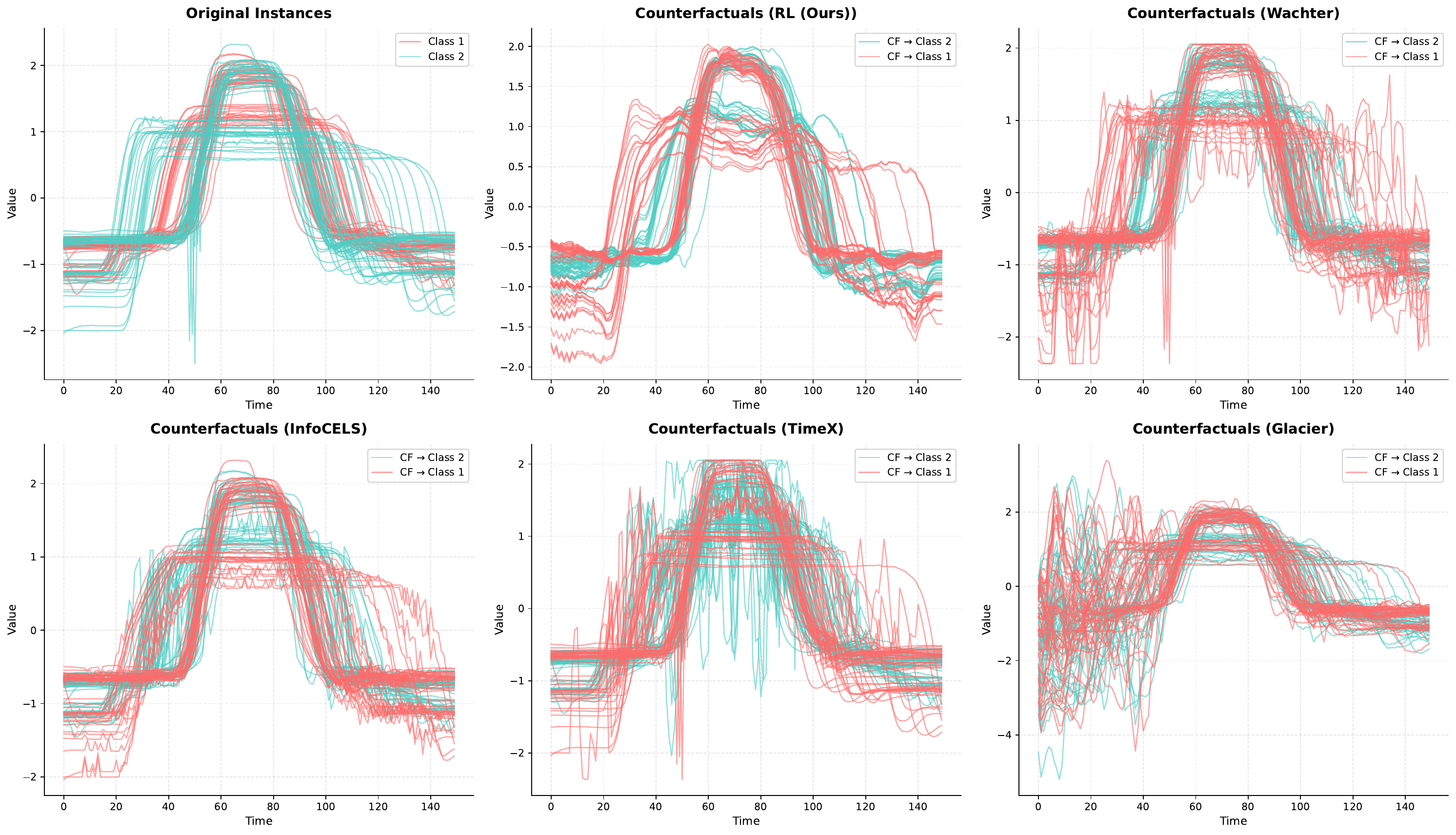}  
    \caption{Qualitative Case Studies on the GunPoint Dataset}
    \label{fig:case_study_gunpoint}
\end{figure}
To complement the quantitative evaluation, we conduct qualitative case studies on the UCR GunPoint dataset—to visually assess the temporal coherence and structural plausibility of counterfactuals generated by different methods. In particular, we plot all valid counterfactuals under each baseline and compare them with our proposed RL-generated counterfactuals. The corresponding visualizations are shown in Figure~\ref {fig:case_study_gunpoint}).

As we can see, even instances from the same class exhibit substantial variation in magnitude and slope, making counterfactual generation more challenging. Most baselines, especially Wachter, TimeX, and Glacier, tend to introduce abrupt local distortions, irregular fluctuations, and unstable peak shapes. In contrast, our method better preserves the global rise--fall motion pattern and produces comparatively smoother and more coherent counterfactuals. Although minor irregular distortions still remain, partly due to the AE-based backbone, they are less severe than those observed in the baseline methods, leading to improved global shape consistency under the dataset’s inherent variability.

\section{Conclusion}
In this paper, we proposed a scalable policy learning framework for counterfactual explanation in time series classification. The key idea is to move counterfactual generation from per-instance test-time optimization to an experience-driven learning process, where an actor learns a transferable policy for producing counterfactual latent representations. We formulate this process as a one-step reinforcement learning problem and operate in the latent space of an autoencoder, enabling the model to generate globally coherent perturbations while avoiding direct optimization over raw time steps. Across 13 UCR benchmark datasets, our method achieves high validity and strong plausibility under multiple anomaly-detection-based metrics. The runtime analysis further demonstrates that, once trained, the learned policy generates counterfactuals with only a single forward pass, making it particularly effective for large-scale counterfactual generation. Overall, our findings highlight policy learning as a promising direction for efficient and reusable time series explanation. Future work will investigate more expressive reward functions, tighter temporal constraints, and broader evaluation on multivariate time series domains.

\bibliography{./mybib}

@inproceedings{delaney2021instance,
  title={Instance-based counterfactual explanations for time series classification},
  author={Delaney, Eoin and Greene, Derek and Keane, Mark T},
  booktitle={International Conference on Case-Based Reasoning},
  pages={32--47},
  year={2021},
  organization={Springer}
}

@article{li2022motif,
  title={Motif-guided Time Series Counterfactual Explanations},
  author={Li, Peiyu and Boubrahimi, Soukaina Filali and Hamd, Shah Muhammad},
  journal={arXiv preprint arXiv:2211.04411},
  year={2022}
}

@inproceedings{li2022sg,
  title={SG-CF: Shapelet-Guided Counterfactual Explanation for Time Series Classification},
  author={Li, Peiyu and Bahri, Omar and Boubrahimi, Souka{\"\i}na Filali and Hamdi, Shah Muhammad},
  booktitle={2022 IEEE International Conference on Big Data (Big Data)},
  pages={1564--1569},
  year={2022},
  organization={IEEE}
}

@article{ismail2019deep,
  title={Deep learning for time series classification: a review},
  author={Ismail Fawaz, Hassan and Forestier, Germain and Weber, Jonathan and Idoumghar, Lhassane and Muller, Pierre-Alain},
  journal={Data mining and knowledge discovery},
  volume={33},
  number={4},
  pages={917--963},
  year={2019},
  publisher={Springer}
}

@inproceedings{filali2022mining,
  title={On the Mining of Time Series Data Counterfactual Explanations using Barycenters},
  author={Filali Boubrahimi, Souka{\"\i}na and Hamdi, Shah Muhammad},
  booktitle={Proceedings of the 31st ACM International Conference on Information \& Knowledge Management},
  pages={3943--3947},
  year={2022}
}

@article{bahri2022shapelet,
  title={Shapelet-Based Counterfactual Explanations for Multivariate Time Series},
  author={Bahri, Omar and Boubrahimi, Soukaina Filali and Hamdi, Shah Muhammad},
  journal={arXiv preprint arXiv:2208.10462},
  year={2022}
}

@inproceedings{bahri2022temporal,
  title={Temporal Rule-Based Counterfactual Explanations for Multivariate Time Series},
  author={Bahri, Omar and Li, Peiyu and Boubrahimi, Soukaina Filali and Hamdi, Shah Muhammad},
  booktitle={2022 21st IEEE International Conference on Machine Learning and Applications (ICMLA)},
  pages={1244--1249},
  year={2022},
  organization={IEEE}
}

@article{wachter2017counterfactual,
  title={Counterfactual explanations without opening the black box: Automated decisions and the GDPR},
  author={Wachter, Sandra and Mittelstadt, Brent and Russell, Chris},
  journal={Harv. JL \& Tech.},
  volume={31},
  pages={841},
  year={2017},
  publisher={HeinOnline}
}

@article{dau2019ucr,
  title={The UCR time series archive},
  author={Dau, Hoang Anh and Bagnall, Anthony and Kamgar, Kaveh and Yeh, Chin-Chia Michael and Zhu, Yan and Gharghabi, Shaghayegh and Ratanamahatana, Chotirat Ann and Keogh, Eamonn},
  journal={IEEE/CAA Journal of Automatica Sinica},
  volume={6},
  number={6},
  pages={1293--1305},
  year={2019},
  publisher={IEEE}
}

@inproceedings{li2023cels,
  title={CELS: Counterfactual Explanations for Time Series Data via Learned Saliency Maps},
  author={Li, Peiyu and Bahri, Omar and Boubrahimi, Souka{\"\i}na Filali and Hamdi, Shah Muhammad},
  booktitle={2023 IEEE International Conference on Big Data (BigData)},
  pages={718--727},
  year={2023},
  organization={IEEE}
}

@inproceedings{breunig2000lof,
  title={LOF: identifying density-based local outliers},
  author={Breunig, Markus M and Kriegel, Hans-Peter and Ng, Raymond T and Sander, J{\"o}rg},
  booktitle={Proceedings of the 2000 ACM SIGMOD international conference on Management of data},
  pages={93--104},
  year={2000}
}

@inproceedings{liu2008isolation,
  title={Isolation forest},
  author={Liu, Fei Tony and Ting, Kai Ming and Zhou, Zhi-Hua},
  booktitle={2008 eighth ieee international conference on data mining},
  pages={413--422},
  year={2008},
  organization={IEEE}
}

@article{scholkopf2001estimating,
  title={Estimating the support of a high-dimensional distribution},
  author={Sch{\"o}lkopf, Bernhard and Platt, John C and Shawe-Taylor, John and Smola, Alex J and Williamson, Robert C},
  journal={Neural computation},
  volume={13},
  number={7},
  pages={1443--1471},
  year={2001},
  publisher={MIT Press One Rogers Street, Cambridge, MA 02142-1209, USA journals-info~…}
}

@article{wang2024glacier,
  title={Glacier: guided locally constrained counterfactual explanations for time series classification},
  author={Wang, Zhendong and Samsten, Isak and Miliou, Ioanna and Mochaourab, Rami and Papapetrou, Panagiotis},
  journal={Machine Learning},
  pages={1--31},
  year={2024},
  publisher={Springer}
}

@inproceedings{sivill2022limesegment,
  title={Limesegment: Meaningful, realistic time series explanations},
  author={Sivill, Torty and Flach, Peter},
  booktitle={International Conference on Artificial Intelligence and Statistics},
  pages={3418--3433},
  year={2022},
  organization={PMLR}
}

@article{li2025info,
  title={Info-CELS: Informative Saliency Map-Guided Counterfactual Explanation for Time Series Classification},
  author={Li, Peiyu and Bahri, Omar and Hosseinzadeh, Pouya and Boubrahimi, Souka{\"\i}na Filali and Hamdi, Shah Muhammad},
  journal={Electronics},
  volume={14},
  number={7},
  pages={1311},
  year={2025},
  publisher={MDPI}
}

@article{dwivedi2023explainable,
  title={Explainable AI (XAI): Core ideas, techniques, and solutions},
  author={Dwivedi, Rudresh and Dave, Devam and Naik, Het and Singhal, Smiti and Omer, Rana and Patel, Pankesh and Qian, Bin and Wen, Zhenyu and Shah, Tejal and Morgan, Graham and others},
  journal={ACM Computing Surveys},
  volume={55},
  number={9},
  pages={1--33},
  year={2023},
  publisher={ACM New York, NY}
}

@article{samoilescu2021model,
  title={Model-agnostic and scalable counterfactual explanations via reinforcement learning},
  author={Samoilescu, Robert-Florian and Van Looveren, Arnaud and Klaise, Janis},
  journal={arXiv preprint arXiv:2106.02597},
  year={2021}
}

@article{lillicrap2015continuous,
  title={Continuous control with deep reinforcement learning},
  author={Lillicrap, Timothy P and Hunt, Jonathan J and Pritzel, Alexander and Heess, Nicolas and Erez, Tom and Tassa, Yuval and Silver, David and Wierstra, Daan},
  journal={arXiv preprint arXiv:1509.02971},
  year={2015}
}

@inproceedings{hosseinzadeh2024acts,
  title={Acts: Adaptive counterfactual explanations for time series data using barycenters},
  author={Hosseinzadeh, Pouya and Li, Peiyu and Bahri, Omar and Boubrahimi, Soukaina Filali and Hamdi, Shah Muhammad},
  booktitle={2024 IEEE International Conference on Big Data (BigData)},
  pages={1327--1332},
  year={2024},
  organization={IEEE}
}

@inproceedings{chen2022relax,
  title={Relax: Reinforcement learning agent explainer for arbitrary predictive models},
  author={Chen, Ziheng and Silvestri, Fabrizio and Wang, Jia and Zhu, He and Ahn, Hongshik and Tolomei, Gabriele},
  booktitle={Proceedings of the 31st ACM international conference on information \& knowledge management},
  pages={252--261},
  year={2022}
}

@article{nguyen2021counterfactual,
  title={Counterfactual explanation with multi-agent reinforcement learning for drug target prediction},
  author={Nguyen, Tri Minh and Quinn, Thomas P and Nguyen, Thin and Tran, Truyen},
  journal={arXiv preprint arXiv:2103.12983},
  year={2021}
}

@article{loh2022application,
  title={Application of explainable artificial intelligence for healthcare: A systematic review of the last decade (2011--2022)},
  author={Loh, Hui Wen and Ooi, Chui Ping and Seoni, Silvia and Barua, Prabal Datta and Molinari, Filippo and Acharya, U Rajendra},
  journal={Computer methods and programs in biomedicine},
  volume={226},
  pages={107161},
  year={2022},
  publisher={Elsevier}
}

@article{bharati2023review,
  title={A review on explainable artificial intelligence for healthcare: Why, how, and when?},
  author={Bharati, Subrato and Mondal, M Rubaiyat Hossain and Podder, Prajoy},
  journal={IEEE Transactions on Artificial Intelligence},
  volume={5},
  number={4},
  pages={1429--1442},
  year={2023},
  publisher={IEEE}
}

@article{vcernevivciene2024explainable,
  title={Explainable artificial intelligence (XAI) in finance: a systematic literature review},
  author={{\v{C}}ernevi{\v{c}}ien{\.e}, Jurgita and Kaba{\v{s}}inskas, Audrius},
  journal={Artificial Intelligence Review},
  volume={57},
  number={8},
  pages={216},
  year={2024},
  publisher={Springer}
}

@article{li2026counterfactual,
  title={Counterfactual Explanation Bake-Off: A Review and Experimental Evaluation for Time Series Classification},
  author={Li, Peiyu and Bahri, Omar and Boubrahimi, Soukaina Filali and Hamdi, Shah Muhammad},
  journal={Machine Learning},
  volume={115},
  number={5},
  pages={114},
  year={2026},
  publisher={Springer}
}
\bibliographystyle{IEEEtran}
\vspace{12pt}
\color{red}

\end{document}